# STOCK TREND PREDICTION USING NEWS SENTIMENT ANALYSIS

Kalyani Joshi[1], Prof. Bharathi H. N.[2], Prof. Jyothi Rao[3]

[1]Department of Computer Engineering, KJSCE, Mumbai
kalyani.joshi@somaiya.edu

[2]Department of Computer Engineering, KJSCE, Mumbai
hodcomp@somaiya.edu

[3]Department of Computer Engineering, KJSCE, Mumbai
jyothirao@somaiya.edu

## ABSTRACT

*Efficient Market Hypothesis is the popular theory about stock prediction. With its failure much research has been carried in the area of prediction of stocks. This project is about taking non quantifiable data such as financial news articles about a company and predicting its future stock trend with news sentiment classification. Assuming that news articles have impact on stock market, this is an attempt to study relationship between news and stock trend. To show this, we created three different classification models which depict polarity of news articles being positive or negative. Observations show that RF and SVM perform well in all types of testing. Naïve Bayes gives good result but not compared to the other two. Experiments are conducted to evaluate various aspects of the proposed model and encouraging results are obtained in all of the experiments. The accuracy of the prediction model is more than 80% and in comparison with news random labelling with 50% of accuracy; the model has increased the accuracy by 30%.*

## 1. INTRODUCTION

In the finance field, stock market and its trends are extremely volatile in nature. It attracts researchers to capture the volatility and predicting its next moves. Investors and market analysts study the market behaviour and plan their buy or sell strategies accordingly. As stock market produces large amount of data every day, it is very difficult for an individual to consider all the current and past information for predicting future trend of a stock. Mainly there are two methods for forecasting market trends. One is Technical analysis and other is Fundamental analysis. Technical analysis considers past price and volume to predict the future trend where as

Fundamental analysis On the other hand, Fundamental analysis of a business involves analyzing its financial data to get some insights. The efficacy of both technical and fundamental analysis is disputed by the efficient-market hypothesis which states that stock market prices are essentially unpredictable.

This research follows the Fundamental analysis technique to discover future trend of a stock by considering news articles about a company as prime information and tries to classify news as good (positive) and bad (negative). If the news sentiment is positive, there are more chances that the stock price will go up and if the news sentiment is negative, then stock price may go down.

This research is an attempt to build a model that predicts news polarity which may affect changes in stock trends. In other words, check the impact of news articles on stock prices. We are using supervised machine learning as classification and other text mining techniques to check news polarity. And also be able to classify unknown news, which is not used to build a classifier. Three different classification algorithms are implemented to check and improve classification accuracy. We have taken past three years data from Apple Company as stock price and news articles.

## 2. LITERATURE SURVEY

Stock price trend prediction is an active research area, as more accurate predictions are directly related to more returns in stocks. Therefore, in recent years, significant efforts have been put into developing models that can predict for future trend of a specific stock or overall market. Most of the existing techniques make use of the technical indicators. Some of the researchers showed that there is a strong relationship between news article about a company and its stock prices fluctuations. Following is discussion on previous research on sentiment analysis of text data and different classification techniques.

Nagar and Hahsler in their research [1] presented an automated text mining based approach to aggregate news stories from various sources and create a News Corpus. The Corpus is filtered down to relevant sentences and analyzed using Natural Language Processing (NLP) techniques. A sentiment metric, called NewsSentiment, utilizing the count of positive and negative polarity words is proposed as a measure of the sentiment of the overall news corpus. They have used various open source packages and tools to develop the news collection and aggregation engine as well as the sentiment evaluation engine. They also state that the time variation of NewsSentiment shows a very strong correlation with the actual stock price movement.

Yu et al [2] present a text mining based framework to determine the sentiment of news articles and illustrate its impact on energy demand. News sentiment is quantified and then presented as a time series and compared with fluctuations in energy demand and prices.

J. Bean [3] uses keyword tagging on Twitter feeds about airlines satisfaction to score them for polarity and sentiment. This can provide a quick idea of the sentiment prevailing about airlines and their customer satisfaction ratings. We have used the sentiment detection algorithm based on this research.

This research paper [4] studies how the results of financial forecasting can be improved when news articles with different levels of relevance to the target stock are used simultaneously. They used multiple kernels learning technique for partitioning the information which is extracted from different five categories of news articles based on sectors, sub-sectors, industries etc.

News articles are divided into the five categories of relevance to a targeted stock, its sub industry, industry, group industry and sector while separate kernels are employed to analyze each one. The experimental results show that the simultaneous usage of five news categories improves the prediction performance in comparison with methods based on a lower number of news categories. The findings have shown that the highest prediction accuracy and return per trade were achieved for MKL when all five categories of news were utilized with two separate kernels of the polynomial and Gaussian types used for each news category.

## 3. METHODOLOGY

### 3.1. System Design

Following system design is proposed in this project to classify news articles for generating stock trend signal.

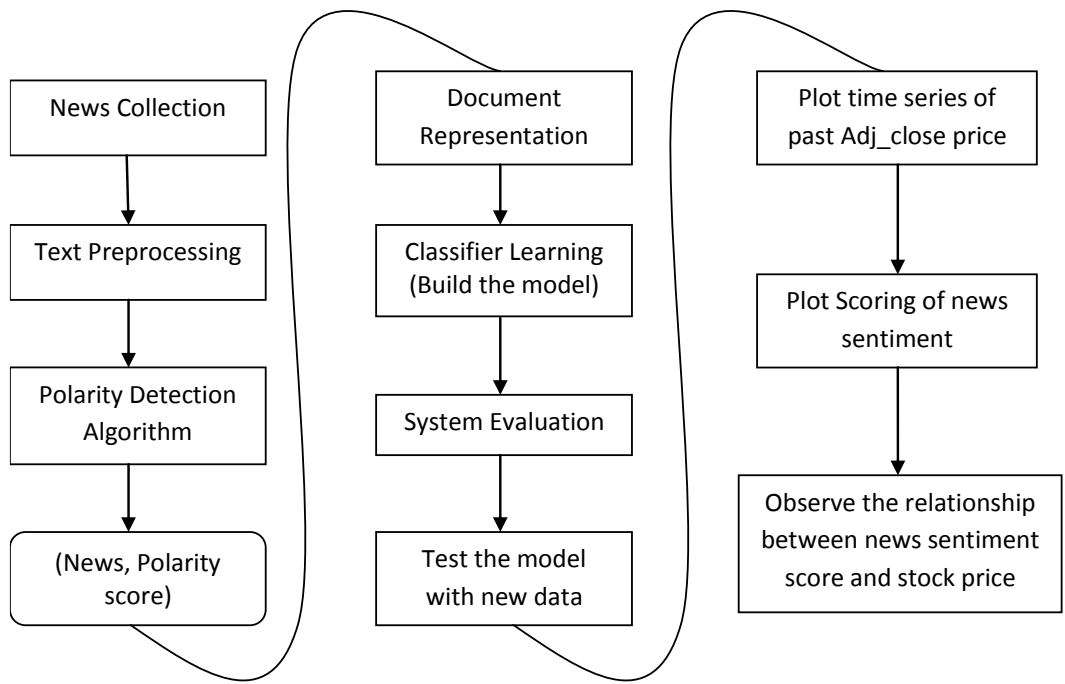

Figure 1: System Design

This design can logically be seen as three phases with first column of blocks in phase 1, second column as phase 2 and third column contains blocks in phase 3. Result of phase 1 is news articles with its polarity score. This result is given as an input to the phase 2. In phase 2, text is converted in tf-idf vector space so that it can be given to the classifier. Then three different classifiers are programmed for the same data to compare results. At the end of phase 2, we evaluate the results given by all classifiers and also test for checking classifier performance for new news articles. In phase 3, we check for relationship between news articles and stock price data. We plot both the data using R language and record the results. In the following sections, each block of the design is explained.

**3.1.1. News Collection**

We collected Apple Inc. Company's data for past three years, from 1 Feb 2013 to 2 April 2016. This data includes major key events news articles of the company and also daily stock prices of AAPL for the same time period. Daily stock prices contain six values as Open, High, Low, Close, Adjusted Close, and Volume. For integrity throughout the project, we considered Adjusted Close price as everyday stock price. We have collected this data from major news aggregators such as news.google.com, reauters.com, finance.yahoo.com.

### 3.1.2. Pre Processing

Text data is unstructured data. So, we cannot provide raw test data to classifier as an input. Firstly, we need to tokenize the document into words to operate on word level. Text data contains more noisy words which are not contributing towards classification. So, we need to drop those words. In addition, text data may contain numbers, more white spaces, tabs, punctuation characters, stop words etc. We also need to clean data by removing all those words. For this purpose, we created own stop-word list which specifically contains stopwords related to finance world and also general English stop words. We built this using reference from [16]. This stop words list contains general words including Generic, names, Date and numbers, Geographic, Currencies.

Also, to ignore words that appear in only one or two documents, we are considering minimum document frequency which considers words that appear in minimum three documents. Stemming is also important to reduce redundancy in words. Using stemming process, all the words are replaced by its original version of word. For example, the words 'developed', 'development', 'developing' are reduced to its stem word 'develop'. Some of the pre-processing is done before applying polarity detection algorithm. And some of them are applied after applying polarity detection algorithm.

### 3.1.3. Sentiment Detection Algorithm

For automatic sentiment detection of news articles, we are following Dictionary based approach which uses Bag of Word technique for text mining. This method is based on the research of J. Bean in his implementation of Twitter sentiment analysis for airline companies [6]. To build the polarity dictionary, we need two types of words collection; i.e. positive words and negative words. Then we can match the article's words against both these words list and count numbers of words appears in both the dictionaries and calculate the score of that document.

We created the polarity words dictionary using general words with positive and negative polarity. Also addition to this, we used Finance specific words with its polarity using McDonald's research [16]. In this dictionary, we collected 2360 positive words and 7383 negative words.

For the news article, we are considering the string which contains headline and news body, both. The algorithm to calculate sentiment score of a document is given below.

Algorithm:

    1. Tokenize the document into word vector.

    2. Prepare the dictionary which contains words with its polarity (positive or negative)

3. Check against each word weather it matches with one of the word from positive word dictionary or negative words dictionary.

4. Count number of words belongs to positive and negative polarity.

5. Calculate Score of document = count (pos.matches) – count (neg.matches)

6. If the Score is 0 or more, we consider the document is positive or else, negative.

Here, we are considering one assumption as if the score of the document is 0, then we label it as positive as we are considering two class problem for this implementation. As a result, we get news collection with its sentiment score and polarity as positive or negative.

### 3.1.4. Document Representation

In order to reduce the complexity of text documents and make them easier to work with, the documents has to be transformed from the full text version to a document vector which describes the contents of the document. To represent text documents, we are using TF-IDF scheme. The higher tf-idf value a term gets, the more important it is. A high value is reached when the term frequency in the given document is high and when there are few other documents in the collection containing the given term/feature. This term weighting method tends therefore to filter out common terms by giving them a very low value.

### 3.1.5. Classifier Learning

As most of the research shows that SVM, Random Forest and Naïve Bayes classification algorithms performs good in text classification. So, we are considering all three algorithms to classify the text and check each algorithm's accuracy. We can compare all the results such as accuracy, precision, recall and other model evaluation methods. All three classification algorithms are implemented and tested using Weka tool.

### 3.1.6. System Evaluation

We divided the data into train and test set. Also, we created unknown data set for classifier to check accuracy of classifier against new data. We evaluated all three classifiers performance by checking each one's accuracy, precision, recall, ROC curve area. The results are as given in the next section.

### 3.1.7. Testing with new Data

News articles from Jan 2016 to April 2016 are used as unknown test set. When comparing results of all classifiers, SVM classifier performs well for unknown data. Random Forest algorithm also worked good comparing to naive bayes algorithm.

### 3.1.8. Plotting the values

After classification of unknown data, we plotted the news score chart and compared with historical price chart.

## 4. EVALUATION

We tested the models using different testing options so that we can compare each method against different scenarios. Following are the test options on which we tested our models.

- 5-fold cross validation
- 10-fold cross validation
- 15-fold cross validation
- 70% Data split
- 80% Data split
- New testing data

| Classification Algorithm | Test Options | | | | | |
|---|---|---|---|---|---|---|
| | **Correctly Classified** | 5-Cross Validation | 10-Cross Validation | 15-Cross Validation | 70% Data Split | 80% Data Split |
| | Random Forest | 86.95% | 89.13% | 88.04% | 92.85% | 88.89% |
| | Naïve Bayes | 83% | 81.52% | 83.69% | 89.28% | 88.89% |
| | SVM | 81.52% | 84.78% | 82.60% | 96.42% | 94.44% |
| | **#Correctly Classified** | 5-Cross Validation | 10-Cross Validation | 15-Cross Validation | 70% Data Split | 80% Data Split |
| | Random Forest | 80 / 92 | 82 / 92 | 81 / 92 | 26 / 28 | 16 / 18 |
| | Naïve Bayes | 76 / 92 | 75 / 92 | 77 / 92 | 25 / 28 | 16 / 18 |
| | SVM | 75 / 92 | 78 / 92 | 76 / 92 | 27 / 28 | 17 / 18 |
| | **ROC Area** | 5-Cross Validation | 10-Cross Validation | 15-Cross Validation | 70% Data Split | 80% Data Split |
| | Random Forest | 0.927 | 0.932 | 0.927 | 0.984 | 0.972 |
| | Naïve Bayes | 0.855 | 0.85 | 0.861 | 0.932 | 0.85 |
| | SVM | 0.824 | 0.853 | 0.834 | 0.971 | 0.958 |
| | **Precision** | 5-Cross Validation | 10-Cross Validation | 15-Cross Validation | 70% Data Split | 80% Data Split |
| | Random Forest | 0.874 | 0.891 | 0.881 | 0.929 | 0.889 |
| | Naïve Bayes | 0.856 | 0.838 | 0.863 | 0.893 | 0.905 |
| | SVM | 0.831 | 0.856 | 0.839 | 0.967 | 0.952 |
| | **Recall** | 5-Cross Validation | 10-Cross Validation | 15-Cross Validation | 70% Data Split | 80% Data Split |
| | Random Forest | 0.87 | 0.891 | 0.88 | 0.929 | 0.889 |
| | Naïve Bayes | 0.826 | 0.815 | 0.837 | 0.893 | 0.889 |
| | SVM | 0.815 | 0.848 | 0.826 | 0.964 | 0.944 |

Figure 2: Comparison of three classifiers against different test options

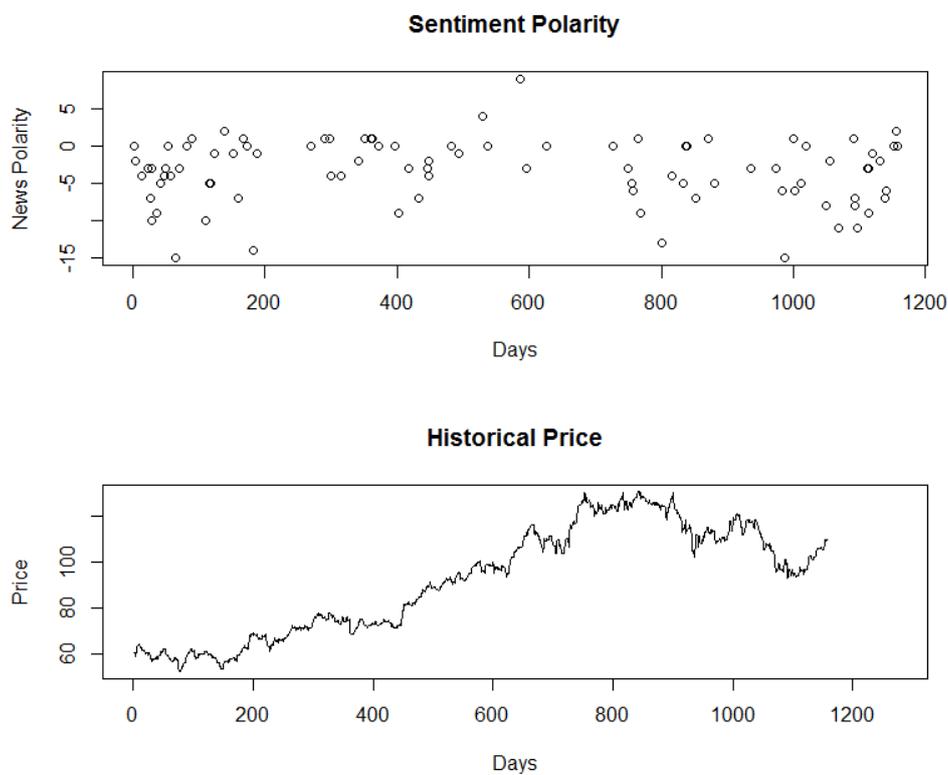

| Correctly Classified | Unknown Dataset |
|---|---|
| Random Forest | 80% |
| Naïve Bayes | 75% |
| SVM | 90% |

| #Correctly Classified | Unknown Dataset |
|---|---|
| Random Forest | 16 / 20 |
| Naïve Bayes | 15 / 20 |
| SVM | 18 / 20 |

Figure 3: Result of testing models with new data

Figure 4: Time series plot of news sentiment score vs. actual stock price for test dataset

## 5. CONCLUSION

Finding future trend for a stock is a crucial task because stock trends depend on number of factors. We assumed that news articles and stock price are related to each other. And, news may have capacity to fluctuate stock trend. So, we thoroughly studied this relationship and concluded that stock trend can be predicted using news articles and previous price history.

As news articles capture sentiment about the current market, we automate this sentiment detection and based on the words in the news articles, we can get an overall news polarity. If the news is positive, then we can state that this news impact is good in the market, so more chances of stock price go high. And if the news is negative, then it may impact the stock price to go down in trend.

We used polarity detection algorithm for initially labelling news and making the train set. For this algorithm, dictionary based approach was used. The dictionaries for positive and negative words are created using general and finance specific sentiment carrying words. Then pre-processing of text data was also a challenging task. We created own dictionary for stop words removal which also includes finance specific stop words. Based on this data, we implemented three classification models and tested under different test scenarios. Then after comparing their results, Random Forest worked very well for all test cases ranging from 88% to 92% accuracy. Accuracy followed by SVM is also considerable around 86%. Naive Bayes algorithm performance is around 83%. Given any news article, it would be possible for the model to arrive on a polarity which would further predict the stock trend.

## FUTURE WORK

We would like to extend this research by adding more company's data and check the prediction accuracy. For those companies where availability of financial news is a challenge, we would be using twitter data for similar analysis. We can also incorporate similar strategies for algorithmic trading.

## ACKNOWLEDGEMENTS

Authors would like to thank our guides, teachers, family and friends who supported in the completion of this research project. Appreciating everyone who helped us knowingly or unknowingly for this project.

**Authors**

Kalyani Joshi

Student of Master in Engineering in at K. J. Somaiya College of Engineering, Mumbai. Completed Bachelors in Engineering from Pune University, 2013.

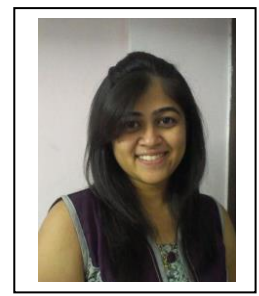

Prof. Bharathi H. N.

Currently working as Head of Department of Computer Engineering at K. J. Somaiya College of Engineering, Mumbai.

Prof. Jyothi M. Rao
Currently working as Associate professor and Associate head of Computer Engineering Department at K. J. Somaiya College of Engineering, Mumbai.